\documentclass{cup-pan}
\usepackage{xcolor}
\usepackage{booktabs}
\usepackage{amsmath}
\usepackage{multirow}

\title{Day-Ahead PV Power Forecasting Based on MSTL-TFT }

\author[1]{Xuetao Jiang}
\author[1]{Meiyu Jiang}
\author[2]{Qingguo Zhou*}

\affil[1]{School of Information Science and Engineering, Lanzhou University, Lanzhou, 730000, China. Email: \url{jiangxt21@lzu.edu.cn} Email: \url{jiangmy21@lzu.edu.cn}}
\affil[2]{School of Information Science and Engineering, Lanzhou University, Lanzhou, 730000, China. Email: \url{zhouqg@lzu.edu.cn}}

\corrauthor{Qingguo Zhou}

\runningauthor{Jiang et al.}

\begin{document}

\maketitle

\begin{abstract}
In recent years, renewable energy resources have accounted for an increasing share of electricity energy.
Among them, photovoltaic (PV) power generation has received broad attention due to its economic and environmental benefits.
Accurate PV generation forecasts can reduce power dispatch from the grid, thus increasing the supplier's profit in the day-ahead electricity market.
The power system of a PV site is affected by solar radiation, PV plant properties and meteorological factors, resulting in uncertainty in its power output.
This study used multiple seasonal-trend decomposition using LOESS (MSTL) and temporal fusion transformer (TFT) to perform day-ahead PV prediction on the desert knowledge Australia solar centre (DKASC) dataset.
We compare the decomposition algorithms (VMD, EEMD and VMD-EEMD) and prediction models (BP, LSTM and XGBoost, etc.) which are commonly used in PV prediction presently.
The results show that the MSTL-TFT method is more accurate than the aforementioned methods, which have noticeable improvement compared to other recent day-ahead PV predictions on desert knowledge Australia solar centre (DKASC).

\keywords{PV forecasting, Deep Learning, Transformer, DKASC.}
\end{abstract}

\section{Introduction}

\begin{table}[htb!]
    \begin{tabular}{ll}
    \toprule
    Glossary &  \\
    \midrule
    PV       & Photovoltaic                                      \\
    ANN      & Artificial neural network \\
    BP      & Back propagation \\
    CNN      & Convolutional neural network \\
    LSTM     & Long short term memory                            \\
    TFT      & Temporal fusion transformer                       \\
    GRN      & Gated residual network                            \\
    GLU      & Gated linear unit                                 \\
    VSN      & Variable selection network                        \\
    NWP      & Numerical weather predictions                     \\
    GHI      & Global horizontal irradiance                      \\
    DHI      & Diffuse horizontal irradiance                     \\
    LOESS    & Locally Weighted Scatterplot Smoothing \\
    MSTL     & Multiple seasonal-trend decomposition using LOESS \\
    DKASC      & Desert knowledge Australia solar Centre \\
    \bottomrule       
    \end{tabular}
\end{table}

Energy demand is increasing dramatically as global urbanization progresses. 
Conventional energy sources such as coal, oil and natural gas have limited reserves, and emit greenhouse gases when converted into electricity, exacerbating the greenhouse effect \cite{iea_GGE}.
To improve public health, more and more countries are using various policies to support clean energy. 
Clean energy is renewable and produces less pollution \cite{epa_2022}, which generally includes solar, wind and hydro energy. 
Compared to conventional energy sources, inexhaustible clean energy exists in all countries and regions. 
Therefore, the development of clean energy has become a global trend.
Solar energy is a kind of clean energy with very low maintenance costs.
Distributed and modular system make solar energy less prone to large-scale failure.
With the steady progress of solar power generation technology, many countries and regions have installed a large number of photovoltaic (PV) facilities, the global PV power generation continues to grow. 
The total cumulative installed PV capacity reached at least 942 GW \cite{iea_2022} at the end of 2021.
The development of solar energy can optimize the energy structure to build a low-carbon and efficient energy system.

PV power generation is highly correlated with the solar radiation and affected by meteorological factors such as floating dust and cloud cover. 
Since the solar irradiation is strongest at noon during the day, the corresponding PV power generation also reaches its peak. 
At this point, other generation methods need to reduce their output as planned to balance the grid demand. 
When PV power generation deviates from the expected value, the grid needs to redeploy power from other suppliers.
This situation requires additional power generation to be temporarily supplemented by other methods, creating a challenge for grid integration. 
On the other hand, the marginal tariff is five times the day-ahead tariff in many electricity markets, excessive inaccuracies can result in penalties, such as offsetting income if the forecast is exceeded by 10\% \cite{MATHIESEN2013357}. 
PV producers need to provide the most accurate valuation of future power generation to achieve the highest possible economic efficiency. 
Therefore, accurately predicted PV generation is of great importance for grid integration.

To describe the day-ahead PV forecasting problem  accurately, we classify and summarize the research on PV forecasting in recent years.

\subsection{Classification by forecasting process}

Some studies use meteorological data to calculate solar irradiance and then predict future PV power output based on PV power generation model, which are indirect prediction methods \cite{ANTONANZAS201678}. 
The PV prediction proposed by Lorenz et al. \cite{https://doi.org/10.1002/pip.1033} uses a model to obtain site-specific irradiance, and then predicts the hourly PV power output for the next two days based on the relationship between irradiance and historical PV power. 
Pelland et al.\cite{https://doi.org/10.1002/pip.1180} used spatial averaging and Kalman filter method to improve the accuracy of solar irradiance prediction by reducing the instability due to cloud variability. 
The PV prediction method is then used to combine solar irradiance, temperature of the PV system surface and historical PV power data to derive the future PV power generation in next 48 hours. 
Since solar irradiance plays an important role in PV power forecasting and therefore precise prediction of solar irradiance is a must. 
Here the authors \cite{ALONSOMONTESINOS2015387} propose a short-term and medium-term solar irradiance prediction method for solar beam, diffuse and global irradiance in sunny, partly cloudy and overcast conditions. 
Lai et al.\cite{LAI2021114941} proposed a deep time series clustering method that divides global horizontal irradiance (GHI) time series data into multiple clusters and then builds a feature attention deep forecasting neural network for each cluster to predict the hour-ahead GHI in the future. 

The direct prediction method directly predicts the PV power output. 
Generally, PV data include meteorological data, historical PV data and PV system parameters \cite{DAS2018912}. 
The authors \cite{https://doi.org/10.1002/eej.20755} compared two methods, direct prediction and indirect prediction, and summarized the use scenarios of the two prediction methods. 
In the primary stage of PV system operation, the data volume is small, the indirect prediction method should be used until the PV system collects enough data. 
After that, using the direct prediction method, the accuracy of the output power prediction of PV power plants can be guaranteed to the maximum extent.

\subsection{Classification by algorithm}
Models for PV power prediction can be generally classified into four categories: PV performance models, classical statistical models, artificial neural network models, and hybrid models.

PV performance models, also known as physical models,  derives PV power generation mainly from the solar irradiance. So no historical data is required, but detailed parameters of the PV plant need to be obtained to achieve accurate prediction \cite{MAYER2022112772}. 
In most cases , the future solar irradiance is obtained from numerical weather predictions (NWP)\cite{LIMA2016807}.
Many PV performance models use additional data to optimize the results, such as meteorological data like temperature, humidity and wind direction, and extra attributes like PV plant capacity and installation angle, leading to increased complexity of the modeling process. 
However, the most important meteorological data, the solar irradiance, still requires NWP to be operative.
Making PV performance models require the assistance of nearby weather stations.

The statistical models use historical data to predict PV power generation. 
Although artificial neural network (ANN) models are statistical models, their data-fit ability is much higher than other classical statistical models. 
Therefore, we split statistical models into classical statistical models and ANN models. 

Classical statistical models include linear and nonlinear regression models, such as linear regression, auto-regressors, and support vector regression. 
They are simple and have a small number of parameters, but are less capable of fitting complex curves. 
Song et al. \cite{7796490} proposed a PCA-SVM model to improve the accuracy of PV prediction, where PCA was used to extract the main features of the data and used as the input to the SVM model.

ANN models include back propagation networks (BP), convolutional neural networks (CNN), recurrent neural networks (RNN), and long-short memory networks (LSTM). 
They have complicated structures, a large number of parameters, so they are tend to be more capable of fitting complex curves than classical models. 
In the past, the parameters number was the limiting factor for model training and applying. 
But now, with GPU, TPU and other computational acceleration methods, the model parameters number is no longer a problem. 
Therefore, the ANN models are always used to predict PV power generation. 
The authors \cite{ALMONACID2014389} propose an ANN-based method that uses two dynamic models to predict the hour ahead GHI and air temperature, and then integrates the configuration parameters of the PV system to calculate the power output of PV plant. 
One study proposes a PV prediction method based on ANN and ELM models, and verifies prediction affect by the training data size, input variables, and variable order\cite{7387113}. 
AlShafeey et al. \cite{ALSHAFEEY20217601} evaluated two different PV prediction modeling techniques, ANN and multiple regression, for predicting PV output power in the next 24 h. 
The experimental results indicated that ANN had higher coefficient of determination values and lower MAE, MSE, and RMSE values, and the ANN model performed better than the multiple regression model. 
The authors \cite{GAO2019115838} use LSTM to solve the problem of PV power output under different weather conditions, and use discrete grey model to the prediction results of PV power output under undesirable weather conditions such as cloudy and rainy days. 
The results show that the RMSE of the LSTM-based prediction method can reach 4.62\% under the ideal sunny days. 
The authors \cite{9621717} propose two extended CNN based frameworks named multi-headed CNN method and multi-channel CNN method, for predicting PV output power one day-ahead and two days-ahead. 
Experiments show that the proposed two frameworks have better prediction results compared with CNN models. 
The authors \cite{munawar2020framework} evaluate the combination of short-term solar power forecasting, and the result show that the combination of feature selection using PCA method and model using XGBoost works better. 
Santos et al. \cite{en15145232} used temporal fusion transformer (TFT) to predict PV data for two regions, Germany and Australia, and achieved RMSE of 6.1\% on the desert knowledge Australia solar centre (DKASC) \cite{dkasc-2008,} dataset.

A hybrid model is a combination of several models by voting, bagging, and bosting, etc.
The combination may contain several different algorithmic models. 
The data they used depend on the models contained inside. 
That is, if two submodels that each require two different kinds of data, the hybrid model would require four kinds of data.
One study \cite{en6020733} proposed a hybrid model containing NWP and statistical models, using statistical models to predict future GHI, together with historical data to predict PV generation. 
The statistical model requires high stability of PV power and meteorological data, and the large deviation of training data will lead to serious errors \cite{AASIM2019758}. 
There are studies in which multiple ANN models are ensembled to achieve improved accuracy. 
Agga et al \cite{AGGA2022107908} proposed a CNN-LSTM model, which uses a CNN to extract features from meteorological data and an LSTM model to fit the output, and fusing the advantages of both models to improve the accuracy of PV prediction. 
The authors \cite{AGGA2021101} propose a hybrid model of CNN-LSTM and ConvLSTM, and concluded that the accuracy of CNN-LSTM and ConvLSTM is higher than the LSTM model. 
Zhen et al \cite{9054985} proposed a hybrid model containing three models, CNN, LSTM and ANN, and used it for PV power prediction. 
The paper eliminates some irrelevant noise by building a mapping model of the sky image and GHI, so that the prediction effect can be improved. 
Then the prediction results of the hybrid model are compared with CNN, LSTM and ANN models. The results show that the hybrid model has better performance and maintains stability under different weather conditions. 

One review of PV forecasting \cite{ANTONANZAS201678} found that more than half of pappers using machine learning models for modeling, which include ANN, LSTM and CNN. 
On the other hand, there are points that good solar prediction methods should involve as much physics as possible \cite{yang2022concise,yang2019guideline}. 
Wolff et al. \cite{WOLFF2016197} compared the prediction performance of support vector regression (SVR) with different inputs and compared the SVR and physical modeling methods, which were similar for a single PV system, but the physical model was better for regional PV prediction. 
The authors \cite{RAMADHAN20211006} compare the accuracy of physical and machine learning models of PV output power and conclude that machine learning models generally outperform physical models if the parameters are chosen appropriately. 
The authors also mention that even though machine learning models have better accuracy, the advantages of physical models with fewer parameters and no training should be taken into account when choosing the optimal model. 
Mayer et al. \cite{MAYER2022112772} compared the predictions of physical models, machine learning models, and hybrid models in the conversion of solar irradiance to PV power, and demonstrated that hybrid models containing the most physically-calculated predictors performed best. 
Hybrid models have a wide range of applications in  predicting the output power of PV generators, and the physical model require lot of input data and detailed requirements, leading to complex modeling \cite{GANDOMAN2018793}. 
Therefore, physical models can be used when predicted meteorological data and PV plant data are available. 
When historical PV and meteorological data are available, statistical models can be used.

\subsection{Classification by forecast horizons}
Forecast horizon is the time span to forecast future PV power generation \cite{DAS2018912}. 
Muhammad et al. \cite{RAZA20151352} published in 2015 classified PV forecasts into three categories, the time horizon from 1 h to 1 week indicates short-term forecasts, from 1 month to 1 year indicates medium-term forecasts, and the time horizon from 1 year to 10 years indicates long-term forecasts. 
Another paper published by Muhammad et al \cite{RAZA2016125} in 2016 classified PV forecasts into four categories, adding ultra-short term forecasts to the original classification, which refers to the prediction of future PV power generation over a time span of 1 min to several min in the future. 
Different prediction ranges have different implications for the power scheduling (generation, transmission and distribution) of PV plants. 
Long-term and medium-term PV forecasts are good for scheduling, but most grid operators only need to make dispatch decisions one day in advance, so short-term PV forecasts directly affect the power consumption plans of PV power stations and are important for the stable operation of power stations \cite{CSEREKLYEI2019358}.

\subsection{PV forecast features}

PV forecast data can generally be categorized into endogenous features and exogenous features. 
The endogenous features include PV generation and the corresponding lagged and logarithmic values, etc. 
Exogenous features include solar irradiance, temperature, humidity, wind direction, and pv plant properties.
Early PV prediction methods were limited by computing power so the data amount is small. 
The used features generally include only solar irradiance and historical power generation \cite{PURI1978409, BACHER20091772}. 
With the development of computers, the variety of parameters for prediction models has increased to include more meteorological features and PV plant properties.

Meteorological features may include global horizontal irradiance (GHI), diffuse horizontal irradiance (DHI), solar zenith angle, temperature, wind speed, humidity, rainfall etc. 
Mellit et al. \cite{MELLIT2010807} used ANN models for grid-connected PV plant prediction, using meteorological features such as temperature, wind speed and cloud density. 
Li et al. \cite{LIN202256} observed that weather type had a significant impact on the fluctuations of PV power: high PV production volume and smooth PV production curve on sunny days; Frequent and dramatic fluctuations curve on cloudy and rainy days, with PV production decreased. 
They classify the weather conditions based on their similarity, and use meteorological data such as solar irradiance and wind direction as additional features.

PV plant properties include azimuth and tilt angle, location, module type, power ratings, efficiency, area, shading effects, tracking effects, manufacturer and aging etc. 
These additional features are important for accurate prediction of PV power \cite{8353805}. 
Akhter et al. \cite{AKHTER2022118185} propose a hybrid deep learning model for hour ahead PV prediction, where three different PV systems are based on polycrystalline silicon, monocrystalline silicon and thin film technologies. 
Based on the availability of PV system parameters such as PV cell temperature and power, Ayompe et al \cite{AYOMPE20104086} compared different models for PV power prediction accuracy and found that the prediction was good using a combination of PV system parameters.

\subsection{Day ahead PV forecasting}

In summary, the short-term forecasts are useful for day-ahead electricity market guidance and power deployment, and most electricity market prices vary with days, making day-ahead PV forecasts a key research direction. 
Among them, ANN-based models are most widely used. 
In addition, the data diversity, data resolution and forecast length are also key parts for studies. 
Therefore, we surveyed the ANN-based day-ahead PV forecasting studies in recent years, as shown in Table 1.

\begin{table}[htb!]
\caption{Day-ahead PV forecasting in recent years}
\resizebox{\textwidth}{!}{%
\begin{tabular}{lllllll}
\toprule
Author &
  Year &
  Model &
  Input Features &
  Resolution &
  Accuracy &
  Data set \\
\midrule
\begin{tabular}[c]{@{}l@{}}Santos et al \cite{en15145232}* \end{tabular} &
  2022 &
  TFT &
  \begin{tabular}[c]{@{}l@{}}PV power, Solar Irradiance, \\ Temperature, Humidity,\\ Solar Zenith Angle, \\ sine/cosine of month\end{tabular} &
  1 h &
  \begin{tabular}[c]{@{}l@{}}NRMSE 6.4\%\\  NMAE 3.3\%\end{tabular} &
  \href{https://dkasolarcentre.com.au/download?location=alice-springs}{DKASC} \\
\begin{tabular}[c]{@{}l@{}}Qu \cite{QU2021120996}*\end{tabular} &
  2021 &
  CNN-LSTM &
  \begin{tabular}[c]{@{}l@{}}PV power,Solar Irradiance, \\ Temperature, Humidity, \\ Daily rainfall, Wind Direction\end{tabular} &
  5-min &
  \begin{tabular}[c]{@{}l@{}}NRMSE 6.34\%  \\ NMAE 4.20\%\end{tabular} &
  \href{https://dkasolarcentre.com.au/download?location=alice-springs}{DKASC} \\
\begin{tabular}[c]{@{}l@{}}Gu \cite{GU2021117291} \end{tabular} &
  2021 &
  WOA-LSSVM &
  \begin{tabular}[c]{@{}l@{}}PV power, Solar Irradiance,\\ Wind Speed,Temperature, \\ Humidity\end{tabular} &
  10-min &
  \begin{tabular}[c]{@{}l@{}}NRMSE 2.55\%  \\ NMAE 2.00\%\end{tabular} &
  Private  \\
\begin{tabular}[c]{@{}l@{}}Wang \cite{WANG2020112766}\end{tabular} &
  2020 &
  LSTM-RNN &
  PV power &
  15-min &
  \begin{tabular}[c]{@{}l@{}} NRMSE 6.29\% \\ NMAE 2.78\%\end{tabular} &
  \href{https://www.esrl.noaa.gov}{ESRL} \\
\begin{tabular}[c]{@{}l@{}}Miraftabzadeh \cite{9203481}\end{tabular} &
  2020 &
  LSTM architecture &
  PV power, Temperature, Time &
  10-min &
  NRMSE 3.35\% &
  Private  \\
\begin{tabular}[c]{@{}l@{}}Theocharides \cite{THEOCHARIDES2020115023} \end{tabular} &
  2020 &
  ANN &
  \begin{tabular}[c]{@{}l@{}}PV power, SolarIrradiance, \\ Temperature, Humidity,\\ Wind Speed and Direction,\\ Solar azimuth and elevation angles\end{tabular} &
  1 h &
  NRMSE 6.11\% &
  Private \\
\bottomrule
\end{tabular}%
}
\end{table}

According to the papers we surveyed in recent years, a significant portion of the researchers used their own private PV power generation data, which is difficult to reproduce and compare. 
To compare with our proposed method, two recent studies using the DKASC are collected. 
On the other hand, different studies have different predicted power ratings and units, which cannot be compared using metrics such as MAE and RMSE. 
Therefore, we used NMAE and NRMSE to normalize the statistical metrics in the above studies from 0 to 100\%.

According to our research, recent day-ahead PV forecasting studies mainly include generation data decomposition, additional meteorological features and pv plant properties, improvement and integration of ANN-based models. 
Due to the instability and variability of PV power generation, further research can be conducted on the above-mentioned short-term PV prediction studies.

\begin{itemize}
  \item Using signal analysis methods to decompose historical PV data. 
  Due to the variability of the natural environment, there is periodicity and randomness in the variation of solar energy, resulting in high fluctuation of PV power sequence data. 
  To achieve high prediction accuracy, many studies  using signal analysis methods (e.g. VMD, EMD  and EEMD, etc.) to decompose the PV generation series into several subsequences. 
  However, it is difficult to make out the association between the subsequences and the original series.
  The time series decomposition algorithm has the same capability as the above methods, and the results have similar periodicity and volatility as the original series.

  \item For PV power generation forecasting, there is no feature categorization based on time series characteristics, and no corresponding model. 
  PV generation forecasting is an application problem and therefore the features have different specific properties. 
  For example, the solar angle changes regularly, the orientation of PV plant is fixed, and there are fluctuations when the solar irradiation changes. 
  Based on our survey, no studies have categorized data based on PV generation forecasting scenarios, and lack of corresponding forecasting models.

  \item Recently, there are few studies on day-ahead PV power forecasting. 
  In contrast, many pappers predict the power point by point, and complete the whole day prediction by iterations. 
  The day-ahead PV power forecasting requires predicting multiple power points at once. 
  The day-ahead PV forecasting is more practical than the iterative one, and more difficult to give accurate predictions. 

  \item There is no comparative analysis under different PV forecasting scenarios. 
  Depending on the actual forecasting conditions, there are differences in corresponding methods. 
  For PV site containing multiple plants, total power prediction can be performed either by summing the individual predictions plant by plant, or by directly predicting the total site power. 
  Depending on the weather station nearby, the data at the forecast time may vary. May include meteorological data such as hourly irradiance and temperature, or only the weather type, or even nothing. 
  The papers we found did not discuss these cases in detail.

\end{itemize}

To address the above issue, we propose an MSTL and TFT based day-ahead PV power forecasting method. 
The main contributions of this study are as follows.

\begin{itemize}
  \item The MSTL, a algorithm for time series decomposition, was adopted to process the historical PV power generation data. 
  For PV prediction problems, common decomposition methods are signal analysis algorithms such as VMD, EMD and EEMD. 
  MSTL is a decomposition algorithm designed for time series problems, which could yield subsequences that reflect the periodicity of PV power series. 
  Results show that MSTL outperforms raw data and other decomposition methods when only historical data is used. 

  \item The data are tagged according to their type and time series properties, which is required by TFT. 
  We tag the data based on their type and whether it changes over time. 
  For example, real features and categorical features, time-varying features and static features, etc. 
  Finally, the tagged data are fed to the TFT model for prediction.

  \item We surveyed day-ahead PV forecasting studies in recent years and normalized their results. 
  The results show that our method has significantly improvement over other studies on the DKASC dataset.

  \item For the power generation forecast of PV site, we perform a comparative analysis of possible scenarios. 
  The results show that the TFT model works better than other common models. 
  Using MSTL decomposition in cloudy and rainy days, and raw data in sunny days are better than other methods in corresponding weather. 
  For PV site forecasting, the summation of each PV plant gives better results than the direct forecast. 
  For NAME and NRMSE on the test set, using meteorological data at prediction have lower values than using only historical data.
  
\end{itemize}

The remaining parts are structured as follows: in Chapter 2, we present the data preparation, models and evaluation metrics in the method. 
The data preparation part includes data description, feature analysis and decomposition algorithms used for PV power sequences. 
The model part introduces adopted forecasting model (i.e. LSTM, RNN-LSTM, CNN-LSTM, XGB Regressor and TFT). 
The evaluation metrics section gives the formulas for NRMSE and NMAE. 
Chapter 3 compares the prediction results of TFT and other models, and the result shows that TFT outperforms than other models. 
Next, we perform a comparative analysis of the TFT model for three cases at PV sites, and offer suggestions for practical PV power prediction by MSTL-TFT based on the conclusions.

\section{Materials and Methods}
\subsection{Data preparation}

The data for this study was collected from the DKASC, the largest PV power station in the Southern Hemisphere. 
DKASC is located in Alice Spring in central Australia, which is a tropical desert climate, with sunshine all year round and little rain, which has great solar energy resources. 
Alice Springs has long summers and short winters. 
Between late April and late November, the weather here is normally sunny. 
Especially in August, the highest percentage of sunny days compared to other months, reaching more than ninety percent, almost no rainfall. 
The remaining five months all have cloudy days. 
The month with the highest percentage of cloudy weather compared to other months is February, with an average of half of the time being cloudy or overcast and less rainfall. 
Moreover, Alice Springs has different hours of sunlight throughout the year. 
The shortest sunshine day is June 21, with about 10 hours, and the longest sunshine day is December 22, with 13.5 hours.

\begin{table}[hbt!]
\caption{PV plants details}
\begin{tabular}{llllll}
\toprule 
PV Plant & Manufacturer   & Array Rating & PV   Technology  & Array   Structure & Install Date \\
\midrule 
PV-01    & BPSolar        & 5            & poly-Si          & Fixed             & 2008         \\
PV-02    & Kaneka         & 5.2          & Amorphoussilicon & Fixed             & 2008         \\
PV-03    & Solibro        & 5.8          & CIGS             & Fixed             & 2017         \\
PV-04    & SunPower       & 5            & mono-Si          & Fixed             & 2009         \\
PV-05    & BPSolar        & 5.1          & poly-Si          & Fixed             & 2008         \\
PV-06    & BPSolar        & 5.3          & mono-Si          & Fixed             & 2008         \\
PV-07    & Trina          & 5.4          & mono-Si          & Fixed             & 2009         \\
PV-08    & Kyocera        & 2            & poly-Si          & Fixed             & 2008         \\
PV-09    & BPSolar        & 6.3          & poly-Si          & Fixed             & 2008         \\
PV-10    & SunPower       & 5            & mono-Si          & Fixed             & 2011         \\
PV-11    & Sungrid        & 5            & mono-Si          & Fixed             & 2010         \\
PV-12    & Sungrid        & 4.9          & poly-Si          & Fixed             & 2010         \\
PV-13    & EvergreenSolar & 16.8         & poly-Si          & Fixed             & 2010        \\
\bottomrule 
\end{tabular}
\end{table}

We selected the PV power generation data from 13 PV plants in DKASC for this study, and their parameters and ID are listed in Table 2. 
The data are collected from 2018 to 2020, with 0.72\% of entire records missing, 0.1\% of individual cell missing, and a few apparent outliers. 
We use maximum or minimum to replace the out-of-bounds values, and interpolate the missing values using their adjacent time points. 
For detecting outliers, skewness works well.
For example, PV-01 had almost zero solar irradiance and temperature on 2018-03-17, while the PV generation values behaved normally. 
These outliers are replaced with the average of the last few identical weather conditions at the same time. 
There are many records here as the interval of the original data is 5 minutes.
To reduce data density, we resampled the data and obtain the hourly generated power.

\subsubsection{Adopted features}

In this study, the data contains timestamp, power production, PV plant status and meteorological features.
The interval between each record is one hour. 
PV plant status includes technology, array structure and manufacturer, etc. 
Meteorological data includes global horizontal irradiation, temperature and rainfall, etc. 
In addition, solar angle is important for PV prediction, so we obtained zenith angle and azimuth angle data from Global Monitoring Laboratory (GML) \cite{ESRL-SGC} for the area where DKASC is located. 
In addition, derived data including lagged values of historical generation, sine and cosine values of months, seasons and weather types. 
Generally, weather could be discerned by the daily solar irradiance. 
According to Eq. (1), the index $k_d$ can be calculated from the daily DHI and GHI, and then the weather type can be derived from it. 
Zang et al. \cite{ZANG2020105790} use $k_d$ ranges to derive weather type in the DKASC dataset, including sunny, cloudy, and rainy. 
We have followed their ranges and listed them in Table 3. 
Where $d$ denotes the date and $t$ denotes the hour.

\begin{equation}
  k_d = \frac{\sum_t{DHI_t}}{\sum_t{GHI_t}}
\end{equation}

\begin{table}[htb!]
\caption{Weather types and their $k_d$ ranges}
\centering
\begin{tabular}{lll}
\toprule
Weather index & Weather type     & k range                         \\
\midrule
1             & Sunny            & 0$\leq k_d \leq$0.15    \\
2             & Partially Cloudy & 0.15$\le k_d \leq$0.45 \\
3             & Overcast/Rainy   & 0.45$\le k_d \leq$1   \\
\bottomrule
\end{tabular}%
\end{table}

\begin{table}[htb!]
  \caption{Adopted features in this study}
\begin{tabular}{llll}
\toprule
Group & Name & From & Tag \\
\midrule
\multirow{4}{*}{\begin{tabular}[c]{@{}l@{}}Endogenous \\ Features\end{tabular}} &
  PV Energy &
  PV Site &
  Time Varying Unknown Reals \\ \cline{2-4} 
 &
  PV Energy Lags &
  Derived &
  Time Varying Known   Reals \\ \cline{2-4} 
 &
  \begin{tabular}[c]{@{}l@{}}Sine of Month,\\ Cosine of Month\end{tabular} &
  Derived &
  Time Varying Known   Reals \\ \cline{2-4} 
 &
  Season &
  Derived &
  Time Varying Known Categorical \\\cline{1-4} 
\multirow{3}{*}{\begin{tabular}[c]{@{}l@{}}Meteorology  \\ Features\end{tabular}} &
  \begin{tabular}[c]{@{}l@{}}Global Horizontal Irradiation, \\ Diffuse Horizontal Irradiation, \\ Temperature, Daily Rainfall, \\ Humidity\end{tabular} &
  PV Site &
  \begin{tabular}[c]{@{}l@{}}Time Varying Unknown Reals\\ Time Varying Known Reals\end{tabular} \\ \cline{2-4} 
 &
  Weather &
  Derived &
  \begin{tabular}[c]{@{}l@{}}Time Varying Unknown   Categorical\\ Time Varying known Categorical\end{tabular} \\ \cline{2-4} 
 &
  \begin{tabular}[c]{@{}l@{}}Solar Zenith Angle,\\ Solar Azimuth Angle\end{tabular} &
  GML &
  Time Varying Known   Reals \\ \cline{1-4} 
\multirow{2}{*}{\begin{tabular}[c]{@{}l@{}}PV Plant \\ Properties\end{tabular}} &
  \begin{tabular}[c]{@{}l@{}}Manufacturer, PV Technology, \\ Array Structure\end{tabular} &
  PV Site &
  Static Categorical \\ \cline{2-4} 
 &
  Array Rating, Install Date &
  PV Site &
  Static Reals\\ 
  \bottomrule
\end{tabular}
\end{table}

We divide the features mentioned in this paper into three groups: endogenous features, meteorological features, and PV plant properties. 
Time-varying tags and static tags are added to the features depending on whether the features change in the time series or not. 
Add known and unknown tags depending on whether the features are knowable in the prediction time period or not. 
All features, groups and tags used in this paper are shown in Table 4. They can help to clearly explain the features performance in the time series. 
For example, the PV power output is time varying unknown reals, indicating that it is a real variable that varies with time and is unknowable at prediction. 
Such labels are required in many Seq2Seq models that combine attention mechanisms, such as the TFT model.

Depending on the data type of the feature, we performed different regularization. 
For the category features, we denote them as their respective one-hot codes. 
For the real features, we process them using the min-max normalization according to their corresponding maximum and minimum values. 
The data in from 2018.1.1 to 2019.12.31 are chosen as the training and validation sets, and the 2020 data as the test set. 
The ratio of training set to validation set is 3:1, and the validation set is stochastic selected in proportion from every month in two years. 

\begin{figure}[hbt!]
\centering
\includegraphics[width=1\textwidth]{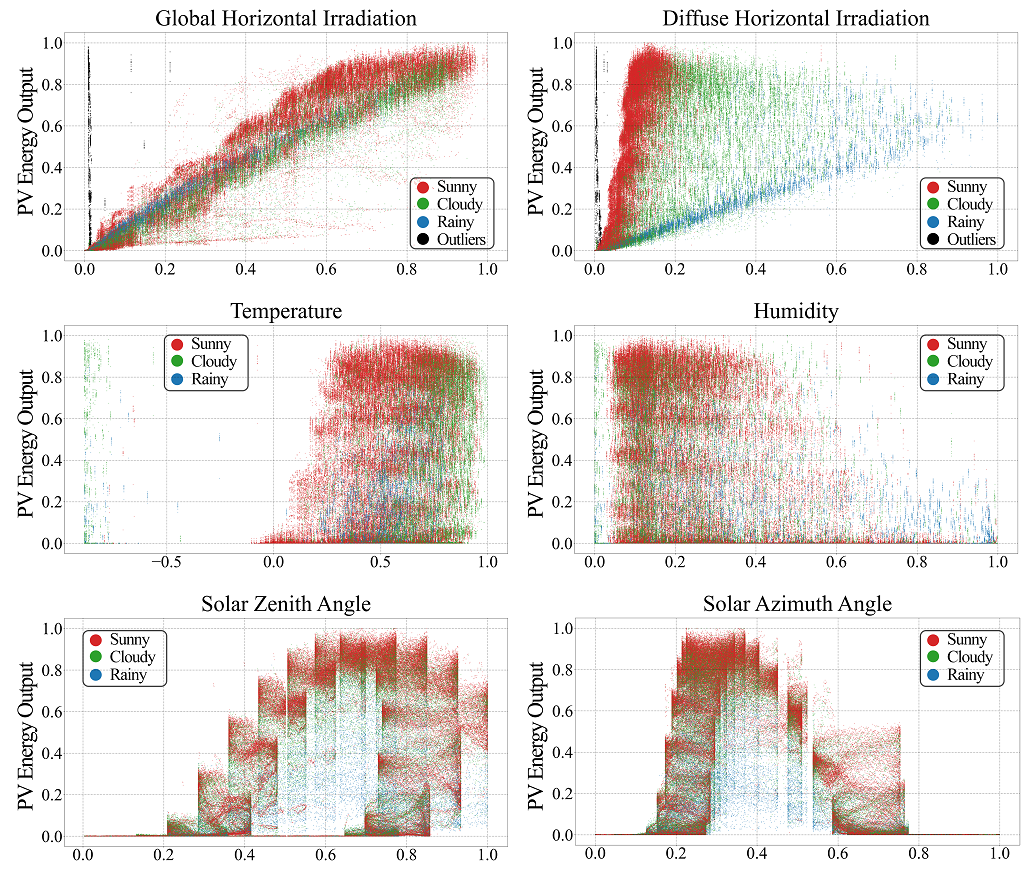}
\caption{Relation between PV energy production and meteorological features (normalized)}
\label{fig:Correlation-Scatters-All}
\end{figure}

Fig. 1 represents the correlation between PV power generation and meteorological features such as GHI, highlighting the differences between weathers. 
The sunny, cloudy and rainy samples are red, green and blue. 
In the first two subplots which representing solar irradiance, black dots indicate outlier points. 
From the figure we see that GHI and DHI have a clear correlation with PV power and a large difference along with weather. 
Especially for the DHI, it is obvious that the data points boundaries are clearer between different weather.

\subsubsection{PV Energy decomposition}

In signal analysis, complex signal sequences can be decomposed to a number of regular subsequences. 
Predicting these subsequences is tended to be easier than predicting the original sequences. 
So many studies use signal analysis decomposition methods to reduce the difficulty of analysis and prediction. 
Common decomposition algorithms for PV power prediction include EMD, EEMD, VMD and their variants. 
In general, the result of the decomposition algorithm includes subsequences and residual, where subsequences represent the derived components, and residual represents the offset between the original value and the sum of all components. 
To clearly compare the results of different decomposition methods, we decompose the PV power sequence by VMD, EEMD and VMD-EEMD, which are common in PV power prediction, and the result are shown in Fig. 2 and Fig. 3.
The data used for the plots are taken from 2018, contains all weather types, five days each.

\begin{figure}[hbt!]
\centering
\includegraphics[width=1\textwidth]{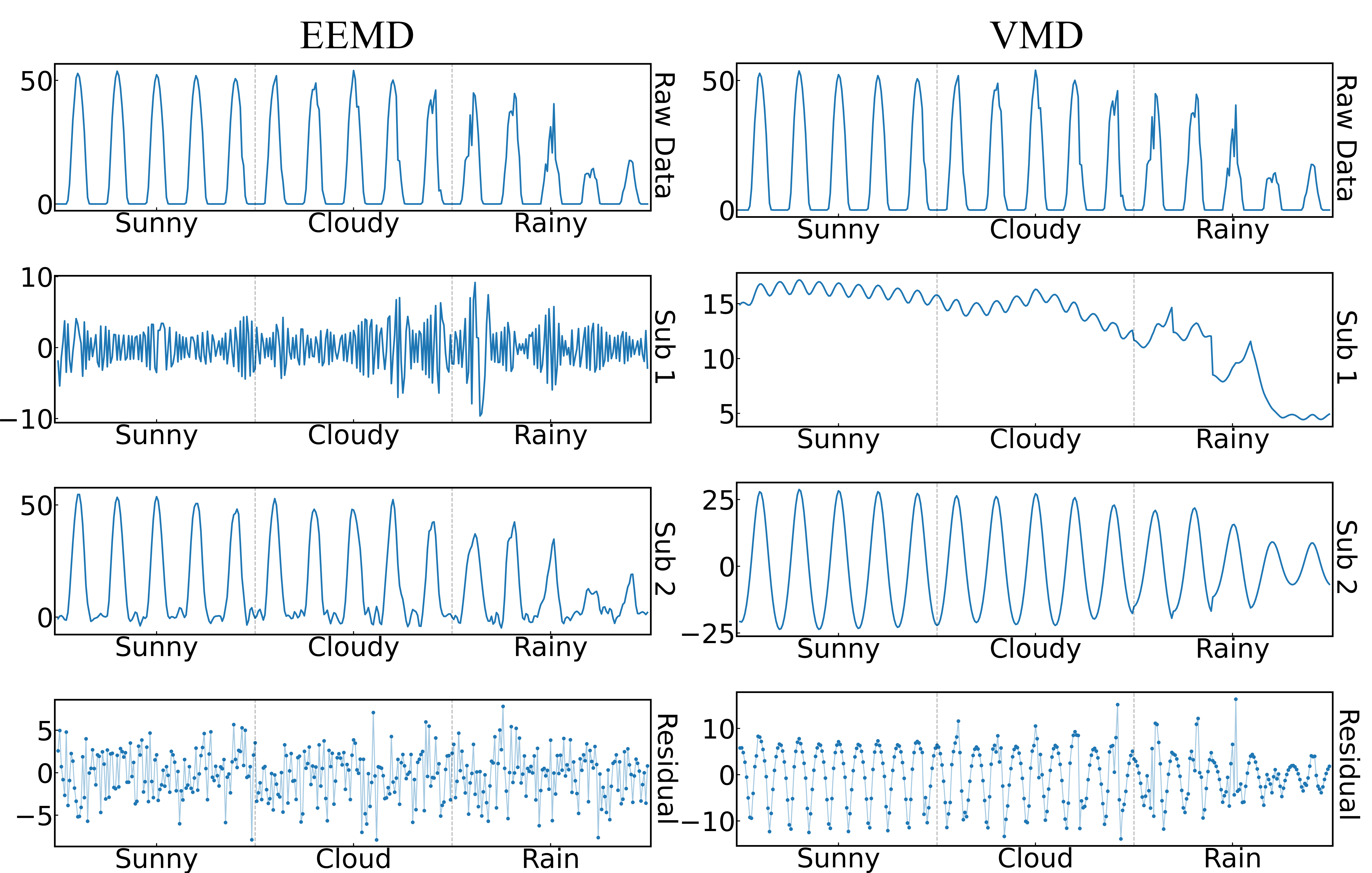}
\caption{Result of EEMD and VMD decomposition}
\label{fig:vmd}
\end{figure}

\begin{figure}[hbt!]
\centering
\includegraphics[width=1\textwidth]{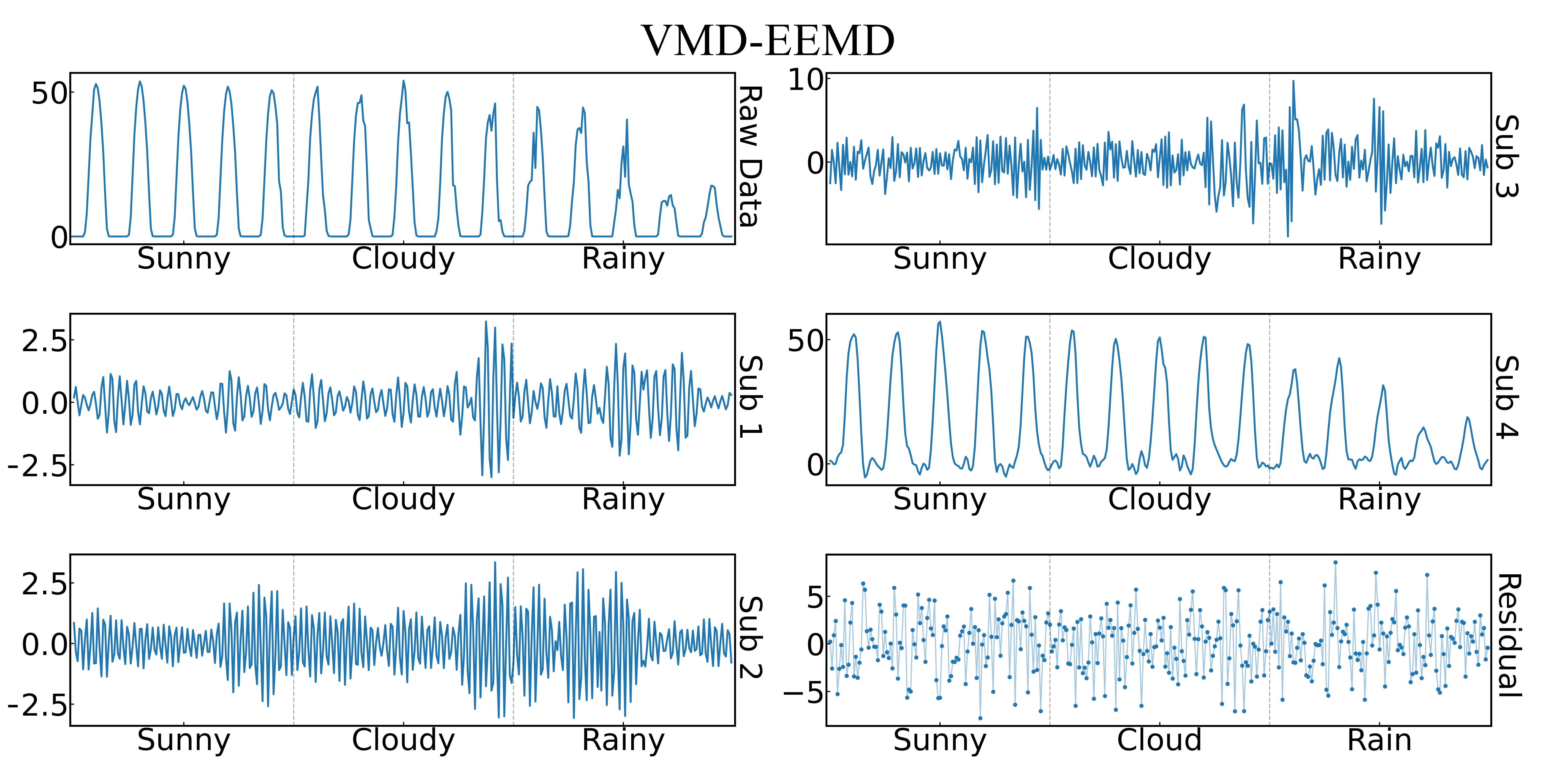}
\caption{Result of VMD-EEMD decomposition}
\label{fig:vmd-eemd}
\end{figure}

From the figures we can see that the PV generation sequence is most regular on sunny days, followed by cloudy days, and is volatile when it rains. 
The results of the decomposition method showed similar trends in the three weathers. 
The frequency of VMD results was moderate, showing periodicity on sunny and cloudy days. 
Some subsequence of EEMD and VMD-EEMD have high frequency. 
Increasing the number of subsequences in EEMD (some studies even reach 10 or more) can reduce the frequency.
However, an excessive number of subsequences may cause some of them to flatten out, and increase the training time for models.

Although the above signal analysis decomposition algorithms are common for PV power forecasting, all of them can hardly reflect the patterns and periodicity of PV power generation over time. 
To tackle this problem, many time decomposition methods have been proposed, such as Prophet \cite{navratil2019decomposition}, STR \cite{dokumentov2015str} and MSTL. 
There are two kinds of time series decomposition, which are additive decomposition and multiplicative decomposition \cite{lu2014analysis}. 
The additive decomposition divides the time series into the sum of trend, seasonality, and residuals. 
The multiplicative decomposition divides the time series into the product of them. 
They can be denoted as Eq. (2) and Eq. (3). 
Where $T_t$ refers to the trend, which indicates the trend of the time series over a longer period. 
$S_t$ refers to the seasonality, which represents the regular fluctuation of the time series in quarterly, monthly, daily, hourly or even shorter periods. 
$R_t$ refers to the residuals, which represent the stochastic fluctuation of the time series, with no obvious trend compared to the two components mentioned above. 

\begin{equation}
  y_t^1 = \hat{T}_t + \hat{S}_t + \hat{R}_t
\end{equation}

\begin{equation}
  y_t^2 = \hat{T}_t \times \hat{S}_t \times \hat{R}_t
\end{equation}

Methods such as STL \cite{cleveland1990stl} and X-13-ARIMA-SEATS \cite{doi:10.1080/07350015.1984.10509398} can perform single seasonal decomposition. 
However, many time series are affected by several periodic factors, which may not be effective for single seasonal decomposition method. 
The author \cite{bandara2021mstl} demonstrate that MSTL has higher accuracy and it is able to model seasonality over time, which requires the lowest computational cost compared to other seasonal decomposition methods. 
MSTL is an extension algorithm based on STL, which can be seen as an iterative use of STL to decompose the multiple seasonal components of a time series. 
The MSTL formula is shown in Eq. (4), where n refers to the number of seasonal components (e.g. quarter, month, day, etc.).

\begin{equation}
  y_t = \hat{S}_t^1 + \hat{S}_t^2 + \cdots +\hat{S}_t^n + \hat{S}_t + \hat{R}_t
\end{equation}

We use the MSTL to decompose the PV power generation sequence, and the result is shown in Fig. 4.

\begin{figure}[hbt!]
\centering
\includegraphics[width=1\textwidth]{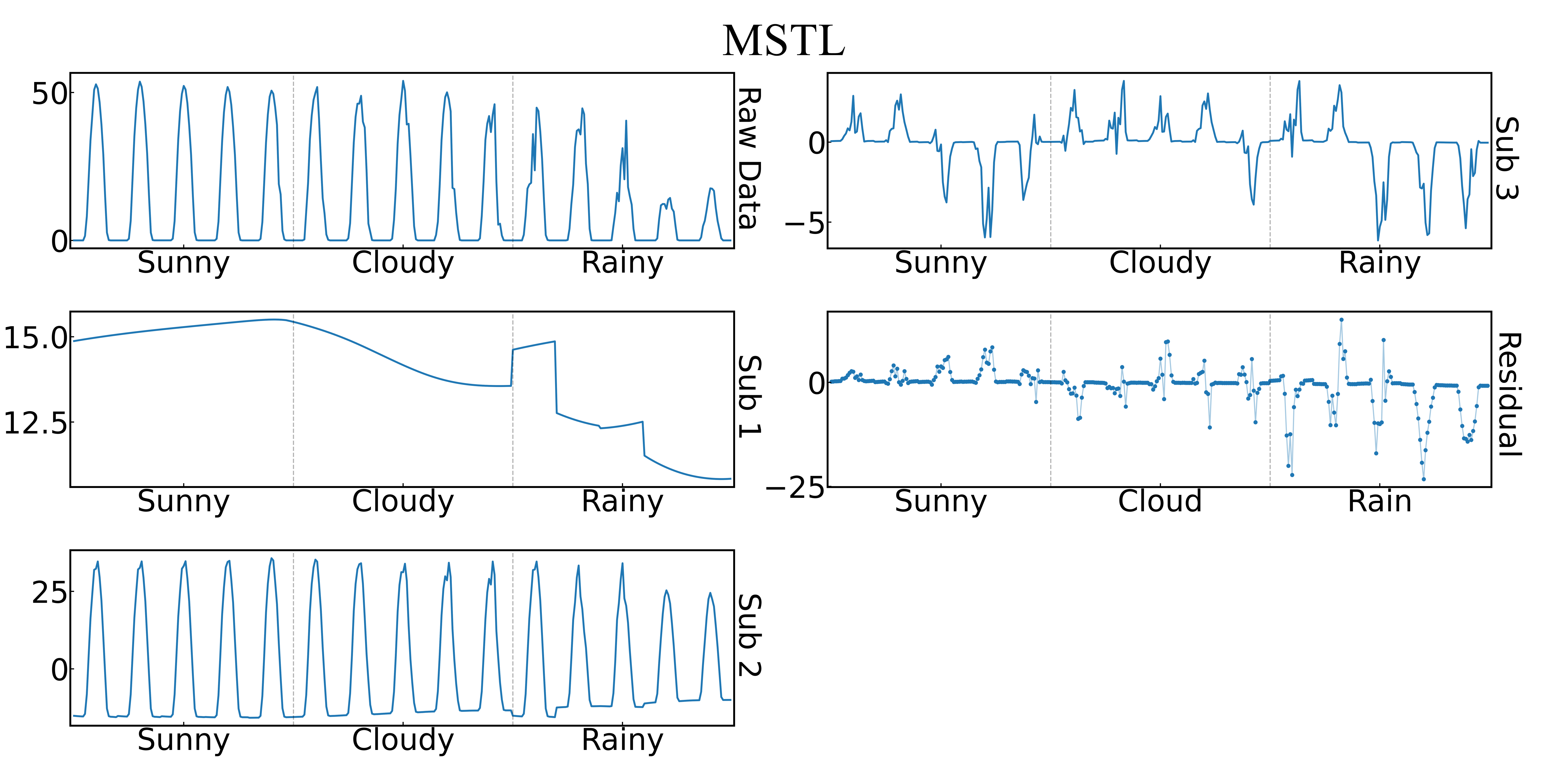}
\caption{Result of MSTL decomposition}
\label{fig:mstl}
\end{figure}

It can be seen that the difference of residuals between MSTL and the other three decomposition algorithms is obvious. 
The distribution pattern of MSTL residuals is closer to human expectations, with small total values and little fluctuations in good weather conditions, which is the opposite situation in poor weather conditions. 
In addition, the frequency of the MSTL subsequence is the same as the original sequence, while all other decomposition methods have higher frequencies than the original ones.

\subsection{Forecasting models}
\subsubsection{LSTM, RNN-LSTM, CNN-LSTM}

We selected four models commonly used in PV power prediction, which are BP, LSTM, RNN-LSTM, and CNN-LSTM. 
In this subsection, we will explain the concepts of these models.
BP model is a kind of classical ANN that refers to a multi-layer feedforward network using the BP algorithm.
It has shown significant improvement in image classification and sequence regression problems compared to other models of the same period.
As a result BP has performed well in early PV forecasts.
For sequence forecasting, RNN have a significant advantage over other ANNs due to their structure of preserving state in sequences. 
RNN model passes the information from the previous time step to the next time step by recurring neurons connections. 
It effectively associates historical information to current input, and is suitable for processing serial data such as time series, text utterances and speech.

With the layers of the network increases rapidly, the error will keep passing forward during the back propagation, resulting in vanishing gradient. 
Due to this problem, RNN models are unable to learn long-term dependencies in sequences. 
LSTM can solve the gradient vanishing by adding additional state units and gate control units on top of RNN, which can perform better in medium and long-term sequence prediction tasks. 
The RNN-LSTM model is an LSTM layer stacked on bottom of an RNN layer, where the RNN receives the input and the result of the LSTM is used as the model output.

CNN is feedforward neural network commonly used to process data with grid topology. 
Time series data and images can be considered as grids of different dimensions. 
CNNs are generally composed of convolutional layers, pooling layers and fully connected layers.
The first two layers are used to extract and downscale features, and the last layers could map features to the model output. 
CNN can integrate high level features and low level features, which can handle the nonlinear relationship between weather data and PV power sequence. 
In CNN-LSTM, the CNN extracts the features and sends them to the LSTM model, and then uses the LSTM model to predict the power output.

\subsubsection{XGB Regressor}

XGBoost is an implementation of boosting in ensemble learning, which forms a strong learner by integrating many tree models together. 
The algorithm grows trees and adds trees by splitting features. 
The tree used is generally CART tree \cite{lewis2000introduction}, including CART classification tree and CART regression tree. 
CART is a binary tree that completes classification or regression by splitting, uses the Gini coefficient to select the optimal features, and prevents overfitting by pruning. 
As an implementation of GBDT, XGBoost optimizes the goal by integrating trees, which has good performance in time series prediction problems \cite{10.1145/2939672.2939785}. 
According to the tree type XGBoost can also be divided into classifier and regressor. 
In the PV power forecasting, the suitable one is XGB Regressor.

\subsubsection{Temporal Fusion Transformer}

TFT uses the attention mechanism in time series forecasting problems, which allows multi-horizon time series forecasting as well as explanatory characterization of time series features. 
This is due to the multi-headed attention mechanism in TFT, which can be evaluated for a time horizon or a number of variables.

Similar to the LSTM that uses gates to control the data flow, TFT uses gated residual network (GRN) and gated linear unit (GLU) to filter non-essential information and avoid undesired non-linear processes. 
GLU borrows the gating mechanism of LSTM and replaces the activation function with the glu function. 
The results are yielded by performing a sigmoid operation on the fully connected network output. 
GRN implements the gate select structure by adding the GLU module, using elu activation function and skip connection.

Variable selection network (VSN) is the key structure of TFT, which derives the weights of input features by GRN.
Then the GRN uses soft-max function to sum up the features as the output. 
And it combines a self-attentive module that uses a multi-headed attention mechanism to integrate the selected features for output, which is used to detect the important features at each step.

TFT applies different data flow operations to static features, time-varying known features, and time-varying unknown features. 
Time-varying features are integrated through VSN, LSTM and gating units, and input to the multi-headed attention mechanism network after GRN. 
Static features are integrated through VSN and static encoders, and the final results are also fed into the multi-headed attention mechanism network.

PV forecasting, as a type of time series forecasting, is essentially a regression problem. 
Common objective functions for solving regression problems are linear objective functions such as mean squared error. 
The variation of PV output power with solar irradiance is non-linear or even non-single trend, as can be seen from Fig. 1. 
At this point, the linear objective functions might show large deviations. 
Therefore, TFT uses the quantile loss, which calculates the difference between the predicted and target values over a certain range. 
The equation for quantile loss is as follows.

\begin{equation}
  QuantileLoss = \sum_{q \in Q} max(q(y_i - \hat{y}_i),(q-1)(y_i - \hat{y}_i))
\end{equation}

Which $y_i$ represent real power value of the i-th point,$\hat{y}_i$ represent corresponding predicted value, $Q$ is quantile values, which is a set like \{0.1,0.25,0.5,0.75.0.9\}.

\subsection{Evaluation Metrics}

In PV forecasting studies, the units of RMSE, MAE can be various (e.g. kW, kWH, etc.), depending on the forecasting goals. 
To visually evaluate our results and to compare with other studies, we used normalized root mean square error (NRMSE) and normalized mean absolute error (NMAE). 
They are in percentages and defined as follows.

\begin{equation}
  NRMSE = \sqrt{\frac{1}{Ty_{max}^2} \sum_{i=1}^T(y_i - \hat{y}_i)^2}
\end{equation}

\begin{equation}
  NMAE = \frac{1}{Ty_{max}}\sum_{i=1}^T{|y_i - \hat{y}_i|}
\end{equation}

where $T$ is prediction length, $y_i$ represent real power value of the i-th point, $\hat{y}_i$ represent corresponding predicted value, and $y_{max}$ is the maximum of real power values.

\section{Result and discussion}

First, PV site generally contains many PV plants, and their total electrical energy constitutes the output of the site. 
For the total output, it can be directly predicted or by the sum of PV plants. 
Second, PV sites obtain weather data from their own weather systems or from nearby weather stations. 
Some sites can obtain future weather data through weather station, while other sites may only have historical weather data. 
Third, the decomposition methods, VMD, EEMD, EEMD-VMD and MSTL, may have different influence on models. 
Therefore, we conducted detailed analysis for three cases mentioned above.

This paper studies day-ahead PV forecasting with a one-day forecast horizon and a three-day input horizon for each forecast. 
The data resolution is 1 hour and the lagged value of PV power generation is 24 hour. 
The 13 PV plants used are named PV-01, PV-02 to PV-13. 
For PV site forecasting, the direct forecast is named by Site-Sum, the summation of each PV plant is named by Site-Indiv. 
When training, all models are deployed with an early stopping strategy to prevent possible underfitting or overfitting due to constant training epochs.

This chapter is structured as follows. 
First, we analyze the performance of BP, LSTM, RNN-LSTM, CNN-LSTM, XGB Regressor and TFT in all cases, shows that TFT is optimal among them. 
Next, three cases are discussed: PV site forecasting, meteorological data availability and PV sequence decomposition. 
Finally, we give some suggestions for the MSTL-TFT model in practical power prediction through the result from this study.

\subsection{Models comparison}

In this subsection, we will compare the prediction loss of BP, LSTM, RNN-LSTM, CNN-LSTM, XGB Regressor and TFT. 
First, we experimented with the output power of the entire PV site and the results are shown in Table 5 and Table 6.

\begin{table}[htb!]
\caption{Models scores when meteorological data is unavailable}
\resizebox{\textwidth}{!}{%
\begin{tabular}{@{}lcccccccccc@{}}
\toprule
       & \multicolumn{5}{c}{BP}                                         & \multicolumn{5}{c}{LSTM}                                        \\ \midrule
Metric & Raw             & VMD    & EEMD    & VMD-EEMD & MSTL            & Raw             & VMD             & EEMD    & VMD-EEMD & MSTL   \\
NMAE   & 7.44\%          & 6.66\% & 9.30\%  & 7.84\%   & \textbf{4.97\%} & \textbf{5.12\%} & 5.54\%          & 5.53\%  & 5.57\%   & 5.59\% \\
NRMSE  & 10.15\%         & 9.20\% & 11.63\% & 10.67\%  & \textbf{7.77\%} & 10.38\%         & \textbf{8.83\%} & 10.33\% & 10.37\%  & 8.92\% \\ \midrule
       & \multicolumn{5}{c}{CNN\_LSTM}                                   & \multicolumn{5}{c}{RNN\_LSTM}                                   \\ \midrule
Metric & Raw             & VMD    & EEMD    & VMD-EEMD & MSTL            & Raw             & VMD             & EEMD    & VMD-EEMD & MSTL   \\
NMAE   & \textbf{3.96\%} & 6.02\% & 5.53\%  & 5.47\%   & 5.03\%          & 5.20\%          & \textbf{4.88\%} & 5.52\%  & 5.50\%   & 5.41\% \\
NRMSE  & \textbf{8.49\%} & 8.95\% & 10.19\% & 10.08\%  & 8.70\%          & 10.45\%         & \textbf{8.14\%} & 10.50\% & 10.73\%  & 8.98\% \\ \midrule
       & \multicolumn{5}{c}{XGBRegressor}                                & \multicolumn{5}{c}{TFT}                                         \\ \midrule
Metric & Raw             & VMD    & EEMD    & VMD-EEMD & MSTL            & Raw             & VMD             & EEMD    & VMD-EEMD & MSTL   \\
NMAE  & 3.62\% & 3.95\% & 5.55\% & 6.72\% & \textbf{3.37\%} & 2.82\% & 2.90\% & 5.47\% & 6.19\% & {\color[HTML]{FE0000} \textbf{2.64\%}} \\
NRMSE & 7.69\% & 6.87\% & 8.62\% & 9.84\% & \textbf{6.21\%} & 7.68\% & 6.25\% & 8.83\% & 9.85\% & {\color[HTML]{FE0000} \textbf{5.63\%}} \\ \bottomrule
\end{tabular}%
}
\end{table}

\begin{table}[hbt!]
\caption{Models scores when meteorological data is available}
\resizebox{\textwidth}{!}{%
\begin{tabular}{@{}lcccccccccc@{}}
  \toprule
  & \multicolumn{5}{c}{BP}                               & \multicolumn{5}{c}{LSTM}                                       \\ \midrule
Metric & Raw    & VMD             & EEMD   & VMD-EEMD & MSTL   & Raw             & VMD             & EEMD   & VMD-EEMD & MSTL   \\
NMAE   & 6.15\% & \textbf{4.74\%} & 8.24\% & 8.98\%   & 5.22\% & 4.82\%          & \textbf{3.90\%} & 4.94\% & 4.79\%   & 4.38\% \\
NRMSE  & 8.16\% & \textbf{6.51\%} & 9.99\% & 10.93\%  & 7.41\% & 8.14\%          & \textbf{5.40\%} & 7.29\% & 7.07\%   & 6.86\% \\ \midrule
  & \multicolumn{5}{c}{CNN\_LSTM}                         & \multicolumn{5}{c}{RNN\_LSTM}                                  \\ \midrule
Metric & Raw    & VMD             & EEMD   & VMD-EEMD & MSTL   & Raw             & VMD             & EEMD   & VMD-EEMD & MSTL   \\
NMAE   & 4.82\% & \textbf{3.59\%} & 5.82\% & 6.90\%   & 3.93\% & \textbf{3.65\%} & 5.43\%          & 3.80\% & 6.02\%   & 4.78\% \\
NRMSE  & 7.42\% & \textbf{5.04\%} & 8.65\% & 10.27\%  & 6.09\% & \textbf{5.92\%} & 7.77\%          & 6.71\% & 11.13\%  & 7.25\% \\ \midrule
  & \multicolumn{5}{c}{XGBRegressor}                      & \multicolumn{5}{c}{TFT}                                        \\ \midrule
Metric & Raw    & VMD             & EEMD   & VMD-EEMD & MSTL   & Raw             & VMD             & EEMD   & VMD-EEMD & MSTL   \\
NMAE  & \textbf{2.40\%} & 3.41\% & 4.49\% & 5.71\% & 2.88\% & {\color[HTML]{FE0000} \textbf{1.04\%}} & 1.59\% & 3.72\% & 4.84\% & 2.04\% \\
NRMSE & \textbf{4.82\%} & 5.49\% & 6.33\% & 7.77\% & 5.27\% & {\color[HTML]{FE0000} \textbf{2.61\%}} & 2.86\% & 4.97\% & 6.30\% & 3.83\% \\ \bottomrule
\end{tabular}%
}
\end{table}

For all models, the scores in Table 6 is better than the scores in Table 5.
Suggesting that the result is better when meteorological data is available.
As can be seen, TFT is the best of them, XGB is the second.
BP, LSTM, RNN-LSTM, and CNN-LSTM have similar NMAE and NRMSE, which has a significant gap from TFT and XGB.
Therefore, for the next evaluation on all PV Plants, we compare only the TFT model, and the results are shown in Table 7 and Table 8.

\begin{table}[hbt!]
\caption{Loss of TFT when meteorological data is unavailable}
\resizebox{\textwidth}{!}{%
\begin{tabular}{@{}lcccccccccc@{}}
\toprule
ID    & \multicolumn{5}{c}{NMAE}                                       & \multicolumn{5}{c}{NRMSE}                                     \\ \midrule
      & Raw             & VMD    & EEMD   & VMD-EEMD & MSTL            & Raw             & VMD    & EEMD   & VMD-EEMD & MSTL           \\ \cline{2-11}
PV-01 & 2.74\%          & 2.87\% & 5.15\% & 6.30\%   & \textbf{2.57\%} & 7.76\% & 6.36\%          & 8.89\% & 9.89\%  & \textbf{5.78\%} \\
PV-02 & 2.67\%          & 2.97\% & 5.30\% & 6.12\%   & \textbf{2.52\%} & 8.20\% & \textbf{6.19\%} & 9.18\% & 10.20\% & 6.53\%          \\
PV-03 & \textbf{2.79\%} & 3.03\% & 5.38\% & 6.38\%   & 2.95\%          & 8.21\% & 6.68\%          & 9.46\% & 10.56\% & \textbf{6.02\%} \\
PV-04 & 2.84\%          & 3.20\% & 5.53\% & 6.72\%   & \textbf{2.76\%} & 8.30\% & \textbf{6.80\%} & 9.38\% & 10.25\% & 6.91\%          \\
PV-05 & 3.12\%          & 3.10\% & 5.47\% & 6.47\%   & \textbf{2.81\%} & 8.52\% & 7.07\%          & 9.79\% & 10.04\% & \textbf{6.93\%} \\
PV-06 & 3.33\%          & 3.23\% & 5.65\% & 6.34\%   & \textbf{2.95\%} & 8.01\% & \textbf{6.50\%} & 8.77\% & 9.89\%  & 6.64\%          \\
PV-07      & 3.07\% & 2.98\% & 5.47\% & 6.45\% & \textbf{2.92\%} & 9.79\% & 9.20\% & 10.73\% & 13.79\% & \textbf{7.35\%} \\
PV-08 & 3.02\%          & 3.30\% & 6.17\% & 9.19\%   & \textbf{2.50\%} & 8.70\% & \textbf{6.37\%} & 9.30\% & 10.35\% & 6.87\%          \\
PV-09 & 3.13\%          & 3.07\% & 5.59\% & 6.57\%   & \textbf{3.03\%} & 8.44\% & 6.64\%          & 9.27\% & 10.24\% & \textbf{6.12\%} \\
PV-10 & 3.01\%          & 3.15\% & 5.52\% & 6.43\%   & \textbf{2.93\%} & 8.32\% & \textbf{5.95\%} & 9.02\% & 10.21\% & 5.99\%          \\
PV-11 & 2.92\%          & 2.89\% & 5.32\% & 6.36\%   & \textbf{2.64\%} & 8.92\% & 6.04\%          & 8.76\% & 9.66\%  & \textbf{5.74\%} \\
PV-12 & 3.37\%          & 3.08\% & 5.26\% & 6.04\%   & \textbf{2.86\%} & 8.16\% & 6.19\%          & 9.09\% & 9.89\%  & \textbf{5.82\%} \\
PV-13 & 2.85\%          & 2.89\% & 5.38\% & 6.18\%   & \textbf{2.62\%} & 7.62\% & 5.82\%          & 7.48\% & 8.02\%  & \textbf{5.50\%} \\
Site-Indiv & 2.59\% & 2.56\% & 3.34\% & 3.81\% & \textbf{2.48\%} & 7.68\% & 6.25\% & 8.83\%  & 9.85\%  & \textbf{5.63\%} \\
Site-Sum   & 2.82\% & 2.90\% & 5.47\% & 6.19\% & \textbf{2.64\%} & 8.14\% & 6.36\% & 8.78\%  & 10.05\% & \textbf{5.72\%} \\ \bottomrule
\end{tabular}%
}
\end{table}

\begin{table}[hbt!]
\caption{Loss of TFT when meteorological data is available}
\resizebox{\textwidth}{!}{%
\begin{tabular}{@{}lcccccccccc@{}}
\toprule
ID    & \multicolumn{5}{c}{NMAE}                              & \multicolumn{5}{c}{NRMSE}                             \\ \midrule
      & Raw             & VMD    & EEMD   & VMD-EEMD & MSTL   & Raw             & VMD    & EEMD   & VMD-EEMD & MSTL   \\ \cline{2-11}
PV-01 & \textbf{0.94\%} & 1.31\% & 3.89\% & 4.97\%   & 1.88\% & \textbf{2.56\%} & 3.34\% & 5.17\% & 6.86\%   & 3.67\% \\
PV-02 & \textbf{1.22\%} & 1.47\% & 3.95\% & 4.68\%   & 2.15\% & \textbf{2.78\%} & 4.42\% & 5.67\% & 6.40\%   & 4.13\% \\
PV-03 & \textbf{1.29\%} & 1.31\% & 3.85\% & 4.82\%   & 2.16\% & \textbf{2.50\%} & 3.97\% & 5.40\% & 6.92\%   & 4.03\% \\
PV-04 & \textbf{0.94\%} & 1.31\% & 3.91\% & 5.27\%   & 2.02\% & \textbf{2.65\%} & 3.53\% & 5.72\% & 7.31\%   & 3.91\% \\
PV-05 & \textbf{1.19\%} & 1.39\% & 4.00\% & 4.93\%   & 2.09\% & \textbf{2.49\%} & 4.10\% & 5.48\% & 6.76\%   & 4.02\% \\
PV-06 & \textbf{1.01\%} & 1.49\% & 3.90\% & 4.97\%   & 2.21\% & \textbf{3.00\%} & 3.87\% & 5.55\% & 6.91\%   & 4.38\% \\
PV-07 & \textbf{0.96\%} & 1.42\% & 3.87\% & 5.09\%   & 2.12\% & \textbf{2.86\%} & 3.45\% & 5.45\% & 6.82\%   & 4.09\% \\
PV-08 & \textbf{1.14\%} & 1.58\% & 4.48\% & 7.33\%   & 2.11\% & \textbf{3.67\%} & 5.15\% & 6.47\% & 9.77\%   & 4.61\% \\
PV-09 & \textbf{1.14\%} & 1.53\% & 3.81\% & 4.77\%   & 1.95\% & \textbf{2.92\%} & 3.68\% & 5.42\% & 6.59\%   & 3.98\% \\
PV-10 & \textbf{1.20\%} & 1.43\% & 3.82\% & 4.83\%   & 2.23\% & \textbf{2.62\%} & 4.08\% & 5.02\% & 6.57\%   & 4.17\% \\
PV-11 & \textbf{1.11\%} & 1.40\% & 3.73\% & 4.92\%   & 2.30\% & \textbf{2.55\%} & 3.16\% & 5.15\% & 6.61\%   & 4.18\% \\
PV-12 & \textbf{1.35\%} & 1.72\% & 3.79\% & 4.65\%   & 2.11\% & \textbf{3.04\%} & 3.39\% & 5.08\% & 6.35\%   & 3.98\% \\
PV-13 & \textbf{1.11\%} & 1.41\% & 3.85\% & 4.77\%   & 2.13\% & \textbf{2.52\%} & 3.24\% & 5.14\% & 6.43\%   & 4.14\% \\
Site-Indiv & \textbf{1.04\%} & 1.01\% & 1.86\% & 2.40\% & 1.78\% & \textbf{2.16\%} & 3.12\% & 3.14\% & 3.93\% & 3.52\% \\
Site-Sum   & \textbf{1.04\%} & 1.59\% & 3.72\% & 4.84\% & 2.04\% & \textbf{2.61\%} & 2.86\% & 4.97\% & 6.30\% & 3.83\% \\ \bottomrule
\end{tabular}%
}
\end{table}

\begin{figure}[hbt!]
\centering
\includegraphics[width=1\textwidth]{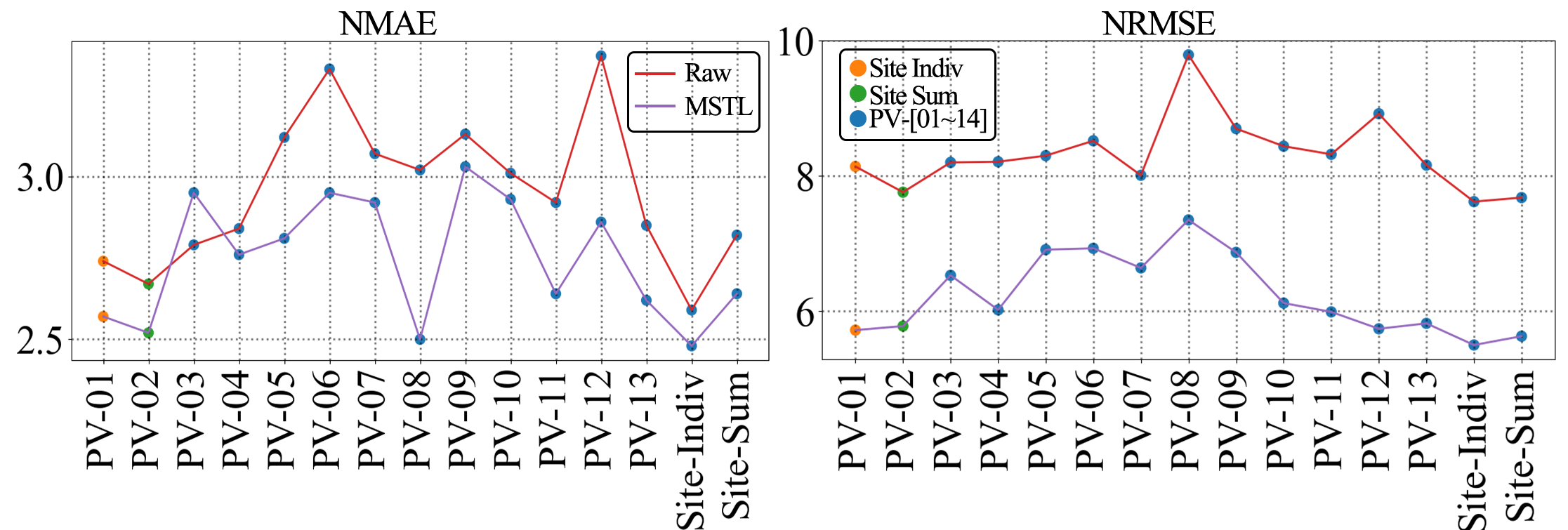}
\caption{Scores of TFT when meteorological data is unavailable}
\label{fig:whcihTFT-Norm}
\end{figure}

\begin{figure}[hbt!]
\centering
\includegraphics[width=1\textwidth]{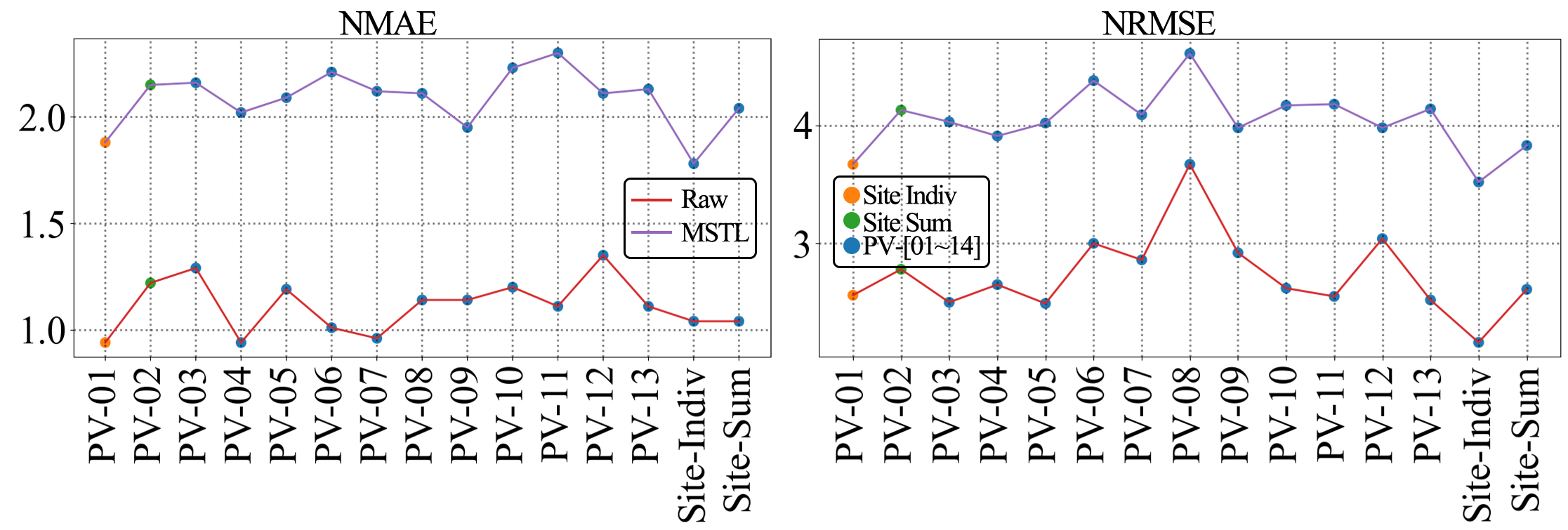}
\caption{Scores of TFT when meteorological data is available}
\label{fig:whcihTFT-NWP}
\end{figure}

In TFT, when the weather data is unknown the MSTL get most minimum values, when the weather data is known more the raw data get most minimum values. 
To show the difference between the two cases above, we have plotted the results for PV site and all PV plants as shown in Fig. 5 and Fig. 6. 
The dashes in the graph indicate the decomposition algorithm (raw data or MSTL), and the scatter indicates the scores of PV plants and PV site.

When meteorological data is unavailable, the MSTL curve is lower than the raw data curve in NRMSE. 
In the NMAE, the curve of MSTL is lower than that of raw data, except for PV-03.
When meteorological data is available, the MSTL curve is lower than the raw data curve in both NAME and NRMSE.

As can be seen, MSTL outperform than other methods when meteorological data unavailable, which have improvement compared to raw data. 
We therefore used MSTL for PV forecasting when meteorological data was unavailable. 
Similarly, we use raw data when meteorological data is available. 
Later we will analyze the performance of TFT in three cases.

\subsection{Case study on TFT}

\subsubsection{Case 1: Individual PV plant and total output}

\begin{figure}[hbt!]
\centering
\includegraphics[width=\textwidth]{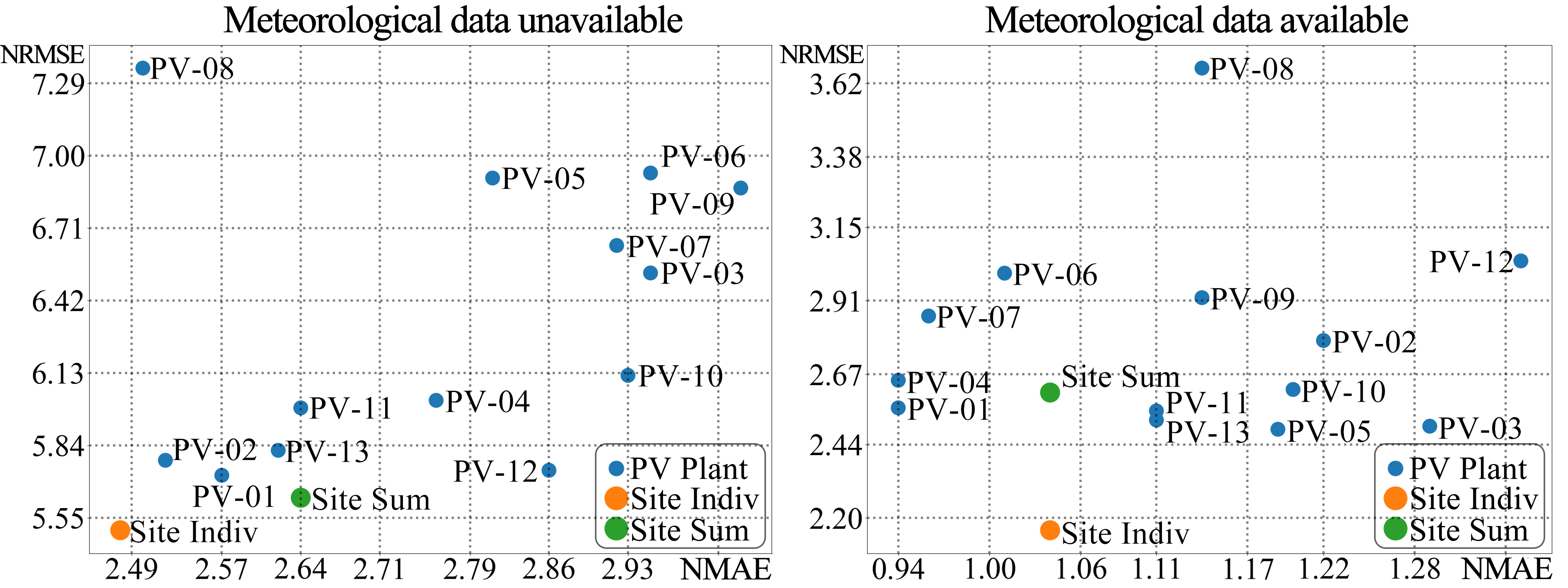}
\caption{Scores of MSTL-TFT when meteorological data is unavailable}
\label{fig:IndivAndSum-Norm}
\end{figure}

In the PV plants forecasting, their point distribution is widely dispersed, and their ranking order is not consistent. 
When predicting the PV site output, we found that the loss for Site-Indiv is lower than Site-Sum, which indicates that the summation of each PV plant gives better results than the direct forecast.
And Site-Indiv's predictions are also better than any other individual PV plant when meteorological unavailable.

Therefore, we conclude that for PV site forecasting, plant-by-plant prediction could give better result than direct forecast, although it is time consuming for model training.
For the individual PV plant forecasts, their score points are scattered and does not show a consistent pattern in both situations. 

\subsubsection{Case 2: Availability of future meteorological data in the prediction}

\begin{figure}[hbt!]
\centering
\includegraphics[width=1\textwidth]{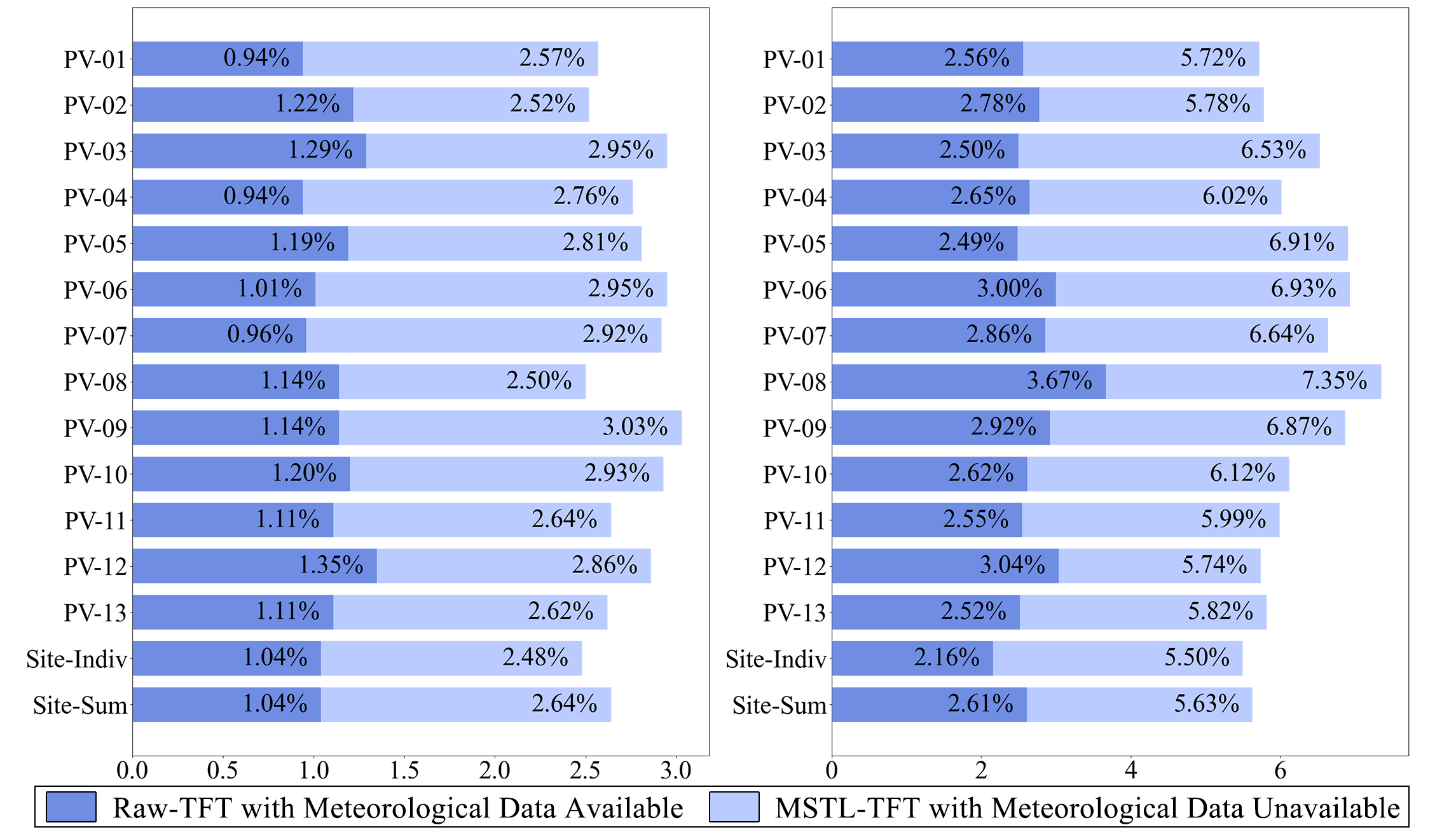}
\caption{Individual PV plant and total output}
\label{fig:Norm-NWP}
\end{figure}

According to the analysis in section 3.1, MSTL outperforms than other decomposition methods when the future meteorological data is unavailable, for raw data is when the meteorological data is available. 
In order to compare the differences of TFT models under different meteorological data conditions, we compare their NMAE and NRMSE respectively. 
The results are shown in Fig. 8.

For PV site and all PV plants, we can see that the NMAE and NRMSE values when meteorological data are available are nearly one-third of the values when meteorological data are not available, which is a significant decrease. 
Therefore, we conclude that when meteorological data are available for the forecast time, the effect is significantly improved over when meteorological data are not available.

\subsubsection{Case 3: PV decomposition on different weather}
To further analyze the impact of data decomposition methods on forecasts, we analyzed the decomposition methods according in different weathers. 
The data contains three types of weather: sunny, cloudy and rainy. 
In the test set, sunny days accounted for 55.40\%, cloudy days accounted for 28.98\% and rainy days accounted for 15.62
\%. 
We counted the losses of each decomposition method under different weather as shown in the following table:

\begin{table}[hbt!]
\caption{Total output when future meteorological is unavailable}
\resizebox{0.8\textwidth}{!}{%
\begin{tabular}{@{}lllllll@{}}
\toprule
          & \multicolumn{3}{l}{NMAE}  & \multicolumn{3}{l}{NRMSE} \\ \midrule
          & Sunny  & Cloudy & Rainy   & Sunny  & Cloudy & Rainy   \\ \cline{1-7}
Raw       & 1.45\% & 3.18\% & 9.37\%  & 4.09\% & 7.72\% & 17.80\% \\
VMD       & 1.61\% & 3.12\% & 8.51\%  & 2.43\% & 5.72\% & 13.05\% \\
EEMD      & 4.11\% & 5.23\% & 12.06\% & 5.28\% & 7.65\% & 19.02\% \\
MSTL      & 1.75\% & 2.87\% & 6.15\%  & 3.29\% & 5.55\% & 11.39\% \\
EEMD\_VMD & 4.79\% & 6.00\% & 13.15\% & 6.14\% & 8.77\% & 20.88\% \\ \bottomrule
\end{tabular}%
}
\end{table}

\begin{table}[hbt!]
\caption{Total output when future meteorological is available}
\resizebox{0.8\textwidth}{!}{%
\begin{tabular}{@{}lllllll@{}}
\toprule
          & \multicolumn{3}{l}{NMAE} & \multicolumn{3}{l}{NRMSE} \\ \midrule
          & Sunny  & Cloudy & Rainy  & Sunny   & Cloudy & Rainy  \\ \cline{1-7}
Raw       & 0.93\% & 1.09\% & 1.34\% & 1.78\%  & 2.56\% & 4.48\% \\
VMD       & 1.05\% & 1.67\% & 3.29\% & 1.58\%  & 2.96\% & 5.22\% \\
EEMD      & 3.58\% & 3.64\% & 3.99\% & 4.54\%  & 4.89\% & 6.08\% \\
MSTL      & 1.61\% & 2.19\% & 3.35\% & 2.81\%  & 4.06\% & 6.02\% \\
EEMD\_VMD & 4.67\% & 4.79\% & 5.39\% & 5.97\%  & 6.24\% & 7.39\% \\ \bottomrule
\end{tabular}%
}
\end{table}

It can be seen from Table (9) and Table (10) that sunny days are better than cloudy days, and cloudy days are better than rainy days. 
This may be due to the regularity of PV generation data on sunny days, which are strongly correlated with time and solar irradiance, and rarely change abruptly. 
On sunny days, the lowest value of NMAE is obtained at raw data, and the lowest value of NRMSE is obtained at VMD. 
On cloudy and rainy days, when future meteorological is unavailable, the lowest values of NMAE and NRMSE are obtained at MSTL, when future meteorological is available, the lowest values of NMAE and NRMSE are obtained at raw data

In sunny days, we conclude that the NMAE values are better when using raw data and NRMSE is better when using VMD. 
In cloudy and rainy days, if future meteorological data is unavailable, then use the MSTL method. 
If future meteorological data is available, then use the raw data.

\section{Conclusion}

For the open-sourced PV power generation dataset DKASC, this paper conducts experiments on four data decomposition methods, VMD, EEMD, EEMD-VMD and MSTL, and five models, TFT, BP, LSTM, RNN-LSTM, CNN-LSTM and XGB Regressor. 
The results show that the TFT model performs the best among the models. 
When the meteorological data in the forecast time is unavailable, the best result of NMAE is 2.48\% and NRMSE is 5.5\%, which are achieved by MSTL-TFT. 
When the meteorological data is available, the best result of NMAE is 1.04\% and NRMSE is 2.61\%, which are achieved by Raw-TFT. 
These results are better than any of the other studies we have surveyed on day-ahead DKASC PV forecasting.

In addition, we analyze the effects of some study cases on TFT PV prediction, which are summarized as follows.

\begin{itemize}
  \item For power forecasting for PV site that contain a lot of PV plants, the summation of each PV plant gives better results than the direct forecast.
  \item For the data decomposition method of the PV generation series, the optimal method depends on the weather conditions and the availability of meteorological data. 
  When meteorological data is available, it is better to use raw data. 
  When meteorological data is not available, then the optimal method is selected based on weather type. 
  For sunny days, using raw data could get lower NMAE values, and using VMD could get lower NRMSE values. 
  For cloudy and rainy days, MSTL makes the best prediction.
\end{itemize}

Based on the results of this paper, we make the following recommendations for the day-ahead PV generation forecast.

\begin{itemize}
  \item In the PV generation forecast, the TFT performs better than other state of the art models such as CNN-LSTM and XGB Regressor.
  \item For PV site forecasting by TFT, the summation of each PV plant gives better results than the direct forecast.
  \item The meteorological data for the forecast time is very important and has a significant improvement compared to no meteorological data. 
  If meteorological data is available, then forecasting should use raw data for forecasting.
  \item Very often, it is not possible to obtain detailed meteorological data for the forecast time, but only the weather type. 
  At this point, the forecast can be based on the next day's weather type. 
  When the weather is sunny, the raw data should be used. 
  When the weather is cloudy and rainy, the MSTL method should be used.
\end{itemize}


\newpage
\bigskip

\paragraph*{Acknowledgments.} 
The authors declare no conflict of interest. The funders had no role in the design of the study; in the collection, analyses, or interpretation of data; in the writing of the manuscript, or in the decision to publish the results.

\paragraph*{Data Availability Statement.} 
Data from DKASC opened data. Code will be opened after receivion.

\printbibliography
\end{document}